# Computoser - rule-based, probability-driven algorithmic music composition

Bozhidar Bozhanov
Independent researcher
bozhidar.bozhanov@gmail.com

**Abstract**

This paper presents the Computoser hybrid probability/rule based algorithm for music composition (http://computoser.com) and provides a reference implementation. It addresses the issues of unpleasantness and lack of variation exhibited by many existing approaches by combining the two methods (basing the parameters of the rules on data obtained from preliminary analysis).

A sample of 500+ musical pieces was analyzed to derive probabilities for musical characteristics and events (e.g. scale, tempo, intervals). The algorithm was constructed to produce musical pieces using the derived probabilities combined with a large set of composition rules, which were obtained and structured after studying established composition practices. Generated pieces were published on the Computoser website where evaluation was performed by listeners. The feedback was positive (58.4% approval), asserting the merits of the undertaken approach.

The paper compares this hybrid approach to other approaches to algorithmic composition and presents a survey of the pleasantness of the resulting music.

## 1 Introduction

Algorithmic composition has been an area of research for a long time and is widely applicable in multiple scenarios, such as elevator music, live accompaniment, aiding human composers.

Computoser was created in order to address five weaknesses of existing algorithms:
- pleasantness;
- similarity between pieces
- memorability of pieces
- requiring supervision or manual input;
- lack a scalable end-user presentation (e.g. a website)

Many algorithms focus on just one of these aspects, which leads to less success in the others. Possible reasons for that are discussed in chapter 7.

## 2 About the algorithm

Computoser is a hybrid, rule-based (knowledge-based), probability-driven (stochastic, per the categories in [1]) algorithm for music composition. It composes by following a set of rules. While progressing with the composition, decisions between multiple allowed alternatives are taken based on predefined probabilities, drawn from both existing musical practice and the analysis of sample data, as described below.

The algorithm cannot be classified as a Markov chain or a Markov decision process, as it does not satisfy the Markov property, due to subsequent choices depending on the result of previous choices.

## 3 Preparation

500 popular songs details were provided by HookTheory.com. Additionally, 50 freely available MIDI files of popular classical and modern composers were analyzed. All entries contained pitch and length of each note. The results were rounded and slightly skewed, and the values presented below are the actual values used in the code, rather than the exact original results.

### 3.1 Music interval and note length probabilities

Table 1 presents the probability for each type of interval to occur. Steps (one note up or down) are the most likely, followed by unisons and skips. Octaves occur rarely, and intervals bigger than an octave even more rarely, and so are excluded from the current version of the algorithm.

| Type | % |
|---|---|
| Unison | 25 |
| Octave | 2 |
| Step | 48 |
| Skip | 25 |

**Table 1** Interval type probability

| Interval | % |
|---|---|
| Perfect 5$^{th}$ | 25 |
| Perfect 4th | 2 |
| Third | 48 |
| Sixth | 25 |

**Table 2** Skip interval probability

The distribution among the skip-type intervals is as shown in Table 2.

The probability of each length (in standard musical terms) is shown in Table 3.

| Note length | % | Note length | % |
|---|---|---|---|
| sixteenth | 10 | dotted quarter | 7 |
| eighth | 31 | half | 9 |
| quarter | 40 | whole | 3 |

Table 3 Note length probability

## 3.2 Rules

The other core aspect is the rule-set. Numerous rules were extracted from multiple sources about music theory and composition [2][3][4][5].

Additional rules were also suggested and existing rules were revised by two composers – Alexeys Pegushevs and Hristo Konstantinov.

The rules can be separated into the following groups:

- Structure – musical structure is at the core of each piece. In this algorithm, the smallest structural component is the motif; multiple motifs form a phrase. The length of each motif varies significantly, within the limits of accepted practice [2].

- Rhythm – note lengths cannot simply be chosen based on the percentages in Table 3 – they must also conform to a rhythm scheme [16]. The algorithm chooses between multiple metres and the length of each subsequent note is fit into the current measure, so that there are no off-beat notes.

- Repetition – according to neuroscience studies [6], repeating structural components (motifs and phrases) makes music memorable. Therefore each component is repeated several times within the piece.

- Variations – simple repetition tends to be "boring" [2], therefore a set of variation techniques need to be employed in a piece.

- Dissonance and syncopation – although music strives to be consonant and rhythmic, unexpected dissonances and syncopation can make it more interesting [7], therefore this element is also encoded in the algorithm

- Endings – endings (candences) determine the completeness of a musical piece.

- Effects – musical effects, including articulation (staccato, tenuto), rubato, background sounds, are employed in order to give a more realistic sounding.

Nearly half of the rules and considerations were selected during the preparation phase and the rest were added later in the process.

All of the rules (the list of which is far from exhaustive), are applied in a given percentage of the cases. Some of these percentages (as shown in sections 3.1 and 3.2) are based on preliminary analysis, and others are based on empirical observations made in the process of writing the algorithm, combined with composition guidelines [8].

## 4 Structure

The structure of the components of the algorithm is simple – there is a list of score manipulators, each of which fills in part of the information for the piece. Manipulators are invoked in a predefined order, as subsequent manipulators depend on the previous ones:

- A part configurer determines what parts will the piece have: pieces always have a main part, and all other parts are optional and selected at random: accompaniment, arpeggio, bass, pads, drone, percussions, simple beats, effects

- A scale configurer determines the scale/scales which the piece is going to be in. The western major and minor scales are the most likely, and scales like Lydian, Dorian, etc. are less likely.

- A meter configurer determines the meter of the piece. For simplicity a piece can have only one metre – e.g. 3/4, 5/8, 6/8, etc.

- A title generator is an extra-musical element, which generates a title for the piece. It is beyond the scope of this paper, except for the fact that adjectives are picked depending on the scale – if it's a minor scale, adjectives are picked from a set of more negative/sad ones.

- Then each part has its own generator. The main part goes first and all other parts follow.

## 5 Algorithm overview

The reference implementation can be seen at: https://github.com/glamdring/computoser

The algorithm is a sequential one, where multiple musical elements are selected while progressing. The selection is informed by a local context, guaranteeing continuity and coherence of the piece.

The main part is at the core of the composed pieces. It contains the melody and most of the rules are applied there. All other parts follow what has already been composed in the main part.

The composition process finishes after a pre-determined amount of measures. The amount of measures in turn depends on the metre and on a random number, picked in a way that it ensures variable length between 1 and 5 minutes. The composition ends with a cadence.

Below are the implementation notes.

## 5.1 Main part component

The implementation of the main part generator is concerned with three main aspects:
- Pitch – what note to play next
- Length – how long should the next note be, in order to fit the predefined meter and to keep a proper rhythm
- Variations – perform different variations on already played motifs.

The "local context" is used to record some of the intermediate decisions. They are later used to determine the subsequent structure / pitches / lengths. This element is violating the Markov property, as pointed out earlier.

### 5.1.1 Pitch

Each subsequent note is chosen according to tables 1 and 2. If the interval to be used is not a perfect fifth or perfect fourth, the interval type (major or minor) that is valid within the current scale is selected. E.g. if a major third would mean a note that does not fall within the notes of the current scale, a minor third is selected.

Table 1 and 2 are used if no additional constraints are required. Such constraints do exist and are:
- There should not be more than two subsequent unstable tones
- Long jumps (more than 7 steps) require a step in the opposite direction
- A predefined sequence may be used – notes from the circle of fifths or part of an expanded chord.

In rare cases (5%) all rules for consonance are disregarded and a dissonant interval (augmented $4^{th}$, minor/major $7^{th}$) is used. This is done (as pointed out in 3.2) for the sake of making the melody more interesting. Too much dissonance, however, would make it unpleasant, so fine-tuning of the probability was needed.

### 5.1.2 Length

Note length is selected according to table 3, unless other constraints need to be taken into account.

Measure size is the primary constraint – all measures should have the same length, as defined by the meter – e.g. 4/4 implies that the equivalent of 4 quarter notes should be used. The last note in a measure is always trimmed in length in order to fit the measure size.

The second constraint is rhythm. Each measure is either simple or compound. Simple measures (2/4, 3/4) have only one down-beat (that is, the moment when the listener would tap their foot). Compound measures have more than one down-beat. For example a 4/4 measure has 2 downbeats – one on the first note, and one on the first note right after the first two quarter notes' equivalent. The algorithm makes sure that on each down-beat a new note is played. A rare exception is the so called syncopation – an already played note is continuing to play at the moment when the down-beat should occur.

Other rules that influence length selection include sequences of same-length notes and not allowing drastic difference in length in subsequent notes – e.g. an eight note cannot be followed directly by a whole note.

### 5.1.3 Variation

Each time a variation of a given motif or a sequence of notes is needed (in a manually fine-tuned percentage of the cases (40%), after at least 2 measures), exactly one of the following standard musical variations is performed:
- Transposition – all notes go higher or lower within the same scale.
- Inversion – the melody is turned "upside-down", adjusting intervals to fall within the same scale.
- Varying the ending – only the final few notes are varied (recursively, using the same method)
- Varying the base structure – only notes on down-beats are preserved, the rest is changed
- Retrograding – the motif is played backwards
- Changing the key – the same melody is played in a different key
- Notes-to-rests – some of the notes (determined at random, using a manually fine-tuned percentage) are replaced by rests (silence)
- Multiply pitches – pitches are multiplied by the same number. They should be kept in scale.

Variations are applied several times (selected at random) to the same motif.

### 5.1.4 The local context

The decisions stored in the local context include: the current direction and contour of the melody, the previously generated phrases, whether to allow more dissonance and syncopation than usual, whether to have ornamentation of notes, whether there is a special sequence of notes to be followed in the next selection (e.g. circle of fifths), what is the current dynamics level, how long should individual motifs be.

## 5.2 Accompaniment part component

The accompaniment part consists of chords. Chords follow the main part, which is already composed separately. Each chord is played in the start of the measure, and is selected to contain a note that harmonizes with the current note in the main part. Chords can be inverted in all possible ways. As with notes, multiple unstable chords cannot follow each other and stable chords are preferred instead.

In the part configurer, it is randomly selected what type of accompaniment is used. Sometimes, instead of a traditional accompaniment, the chords can be arpeggiated – played note-by-note until filling the measure (some notes from the chord may be played more than once)

## 5.3 Other parts' components

Additional parts are sometimes added (each part has separate probability to appear, where the probabilities are manually fine-tuned), for a more varied output:
- Pads – long-sounding background tones
- Simple beat – a simple percussion beat on each first note or down-beat
- Percussions – random (but consistent, with manually fine-tuned percentage) percussion patterns, with strokes more likely to be falling on down-beats.
- Bass – a bass line with a very simple melody based on stable tones from the scale in the main part.
- Drone – a single note played multiple times per measure, throughout all measures.
- Timpani – one of several timpani patterns is played in a small random number of measure, for a more dramatic sounding
- Effects – birds, bells, applause and other effects are played at random moments during the piece.

## 6 Evaluation

There are four stages of evaluation:
1. Post-composition checks – the algorithm verifies if each newly generated piece conforms to basic musical requirements: is it rhythmic (does it have syncopated notes), is it harmonic (does it have notes that are outside the defined scale). Discrepancies are reported in the form:
   Piece X has unbalanced measures;
   Piece X has 9 out-of-scale notes.
2. "Sanity check" performed by a limited audience.
3. Evaluation by human listeners on the website, where new pieces are generated regularly.
4. Based on the human evaluation, an analysis is carried out on the database to investigate what are the least likes pieces and what they have in common. So far the following have been identified as the main reasons for not being liked: particular instruments (likely due to their artificial sound) and rhythmic discrepancies.

There have been 6000 evaluations by users, of which 3500 were positive and 2500 were negative. The total number of generated pieces so far is 8000. They have been played a total of 170000 times by 28000 unique users. The list of "top tracks" indicates that at least a sample of pieces is considered of high musical quality (http://computoser.com/toptracks).

The perceived quality of a piece depends on the quality of MIDI files, which in turn depends on the soundbank in use. Since a free soundbank was used, the lower MIDI quality has influenced the listeners who gave negative feedback. A survey with 44 respondents (http://form.jotformeu.com/form/42482375120348) was conducted and 41% of the respondents selected "artificial sound" as a primary reason for not liking Computoser pieces. (23% selected "the rhythm", 13% "the structure", 12% "the melody" and 11% "the harmony").

The same survey was used to evaluate the pleasantness of a pre-selected generated piece, as it is subjective and can't be determined via formal means. The average response (from 1 to 5) was 2.85, indicating that the music is perceived as generally pleasant.

## 7 Comparison to other approaches

No existing algorithm is deployed live, except for melomics (http://melomics.com). Comparison to other algorithms is hard, due to the subjective nature of music perception, but the survey, comparing sample pieces by 7 popular algorithms, resulted in Computoser sharing the first place with DarwinTunes (which depends on human input). People had to select pieces of which 2 algorithms they liked the most, and the result was: DarwinTunes[11] (42% liked this algorithm), Computoser (42%), SoundHelix [9] (34%) Iamus(18%), Melomics (18%), Wolfram Tones (11%) and Fractal Music [14] (7%)

Below is a short comparison with other approaches to algorithmic music composition, as reviewed in [1], [17].

Purely mathematical approaches, including fractals, mathematical functions and other crude transformations from numbers to pitches have a significant flaw – they cannot be pleasant to listen to, because they do not account for the physical characteristics of sounds and their interactions. The harmonic series, which are at the core of music being perceived as pleasant, do not follow directly from arbitrary mathematical constructs. That's why algorithmic music using solely mathematical approaches produces unpleasant music [14] (as per the survey). Even if it is limited to produce music in a given scale, it still cannot follow composition rules.

Purely rule-based approaches lack variation in what they produce, because the number of possible outcomes of strictly following rules is limited.

Sample recombination is another approach, pioneered long before computers were invented – taking a set of musical motifs and recombining them in multiple possible ways. They rely on preliminary composition and selection of samples, which involves human composers. But more importantly, they also have a limited (albeit big) number of possible compositions, and are often very similar to one another (e.g. the SoundHelix algorithm [9]). We can probably label this approach as "hardcoded algorithmic composition". The approach in Computoser, on the other hand, allows for much more variation in the results.

Evolutionary algorithms are a prominent approach, used in the Iamus algorithm [10]. They initially produce any sequence of sounds, which then evolve until pleasant music is produced. The problem is the fitness function – since music is subjective, there can't be an objective fitness function. Some algorithms, like DarwinTunes, use human evaluation, but that slows down after several thousand generations and doesn't get significantly better [11]. Others use composition rules (as the one presented in this paper) to determine which child is fit [10][12]. While this is a viable approach, it requires way more computation power than using simple randomness, while the results are practically the same – random sequence of notes that follow a set of loosely defined rules.

Purely statistical and machine learning based approaches "train" on a given set of existing musical pieces and then produce new pieces [13][15]. The resulting music is heavily influenced by the input data, e.g. a classical training set would mean that the algorithm produces only classical music. Computoser, on the other hand, is not limited to a particular style, even though its initial analysis was based on pop and classical music.

## 8  Future work

The algorithm implementation is basic and limited, and can be extended and improved by adding more rules for non-main parts, adding more types of parts and refining the main part rules. The sound quality can be improved by carefully selecting a soundbank.

A more thorough analysis can be performed as part of the evaluation (chapter 6, point 4) to look for musical features that are more likable than others.

The algorithm can then be used as a helper tool for composers, giving them particular motifs as well as a continuous stream of background music.

## 9  Conclusion

After studying and comparing different approaches, it can be said that a hybrid probability/rule based approach, where rules are loosely defined, is at least as good as the state-of-the-art. And while evolution of music itself does not seem justified, rules can evolve with human feedback serving as fitness function.

In addition to the approach, the data and rules gathered in the preparation phase may be a valuable resource for future works.

Computoser presents a scalable end-user presentation of a potentially endless amount of musical pieces, generated with no human input, pleasant to listen to, differ significantly from each other and are memorable.

These improvements, measured by a survey, signify the prominence of the selected approach.

## 10  References

[1] Papadopoulos, G & Wiggins, G, AI Methods for Algorithmic Composition: A Survey, a Critical View and Future Prospects, presented at 1999 AISB symposium of musical creativity http://jakeadams.googlecode.com/svn-history/r31/trunk/CS438/HW/Paper/AISB99b.pdf
[2] Vahromeev, VA, Elementary Music Theory (in Bulgarian), 1968, IV edition, Science and Art, Sofia
[3] Jarrett, S & Day, H. Music Composition for Dummies, 2008, Wiley Publishing, Hoboken, NJ, ISBN 978-0-470-22421-2
[4] Koralov, A, Aesthetical Analysis of Musical Pieces (in Bulgarian), 1979, Peoples' Education, Sofia
[5] Music Theory Online, http://www.musictheoryhelp.co.uk/guides/
[6] Levitin, D, This Is Your Brain On Music, 2008, Atlantic Books, ISBN 978 1 84354 716 7
[7] Hoffman, M, 1997, National Symphony Orchestra. NPR http://www.kennedy-center.org/nso/classicalmusiccompanion/syncopation.html
[8] Music & Melody, http://mannycepeda.biz/cbt/lessons.html
[9] Schrger, T, SoundHelix, http://www.soundhelix.com/
[10] Diaz-Jerez, G Composing with Melomics: Delving into the Computational World for Musical Inspiration https://muse.jhu.edu/login?auth=0&type=summary&url=/journals/leonardo_music_journal/v021/21.diaz-jerez.pdf
[11] MacCallum, B et al, Evolution of music by public choice, http://www.pnas.org/content/109/30/12081.abstract DarwinTunes http://darwintunes.org/
[12] Esler, R, Performing algorithmic computer music: real-time score interpretation of David Birchfield's community art: resonant energy http://www.tml.tkk.fi/Studies/Tik-111.080/2000/papers/hanna/alco.pdf
[13] Cope, D, 1996, Experiments in Musical Intelligence, A-R Editions, ISBN: 978-0895793379
[14] Hazard, C & Kimport, C, 1999, Fractal Music, http://www.tursiops.cc/fm/
[15] Chan, M, Potter, J, Schubert, E, Improving algorithmic music composition with machine learning http://homepages.inf.ed.ac.uk/s0786354/publications/icmpc06.pdf
[16] Blackwell, T. et al, Sound and Music, Goldsmiths, University of London, Subject Guide http://www.londoninternational.ac.uk/sites/default/files/computing-samples/co3346_ch1-2.pdf
[17] Fernández, J. D. & Vico, F. (2013) AI Methods in Algorithmic Composition: A Comprehensive Survey. Journal of Artificial Intelligence Research. 48513–582.